# Building Function Approximators on top of Haar Scattering Networks

Fernando Fernandes Neto

*Abstract*—In this article we propose building general-purpose function approximators on top of Haar Scattering Networks. We advocate that this architecture enables a better comprehension of feature extraction, in addition to its implementation simplicity and low computational costs. We show its approximation and feature extraction capabilities in a wide range of different problems, which can be applied on several phenomena in signal processing, system identification, econometrics and other potential fields.

## I. INTRODUCTION

The field of artificial neural networks has exploded during the 1980s due to its universal approximation capabilities, as can be seen in [1], but the lack of understanding of the underlying statistical and geometric features extracted from the analyzed signal discouraged significantly its usage among scientists and researchers, as can be seen in [2-3]. Since then, most of its usage has been relegated to applications where such understanding can be neglected, such as computer vision, non-linear state-space estimators and other tasks related to control where exact algorithmic approaches are unknown or too difficult to implement, according to [3].

More recently, aiming to enlightening these black-boxes, several approaches have been under heavy development, such as variables contributions in the feed forward structure [4], visualization using saliency maps [5], generation of skeletal structures [6], fuzzy rule based evaluation of all permutations [3], extraction of functional relations using sensitivity analysis of input data [7], as many others.

In a parallel way, other researchers have been successfully developing new kinds of feed-forward neural architectures that behave much more like a transparent box, where the extracted features can be directly evaluated and understood. Convolutional Neural Networks are a great example of such achievements, as can be seen in [8-10]. Despite its several layers, they can be employed on different types of tasks, including text classification, natural language processing, computer vision and so on, with a good understanding of what is happening behind the curtains.

Manuscript received January 15, 2018. This work was supported in part by the FIPE (Institute of Economic Research Foundation) by means of a post-doctoral scholarship.

Fernando Fernandes Neto is with the University of São Paulo, São Paulo, Brazil (e-mail: fernando_fernandes_neto@usp.br).

Basically, according to [8], this kind of networks alternate linear operators, whose coefficients can be optimized with training samples and provide extraction of features of the dataset, with pointwise nonlinearities, which usually imposes an invariant transform, allowing dimension reduction of the problem.

Hence, taking into account these recent developments, which are supported over the concept of feed-forward scattering networks, the idea of this paper is to propose a more transparent and general-purpose approximators/classifiers using Haar Scattering Networks.

On the other hand, for the sake of simplicity, we do not intend to prove if this concept can be applied everywhere, nor verify in which conditions it holds. Keeping that in mind, it seems important to define what Haar Scattering Network is and what our approach is to start developing such tool. Afterwards, we show that the features extracted by our Haar Scattering Network provide understandable information from the original signal in a set of different problems. Also it can be easily used by a simple linear classifiers or simple regression structures, enabling us to build simple and powerful general-purpose approximators.

In the last session, we discuss the main results, some possible generalizations of the results for multivariable/multiparameter analysis and point some directions for future research.

Consequently, this paper is divided into the following sections: Introduction; A Brief Review of Wavelets; Description of Haar Scattering Networks; Computational Examples; Discussion of the Results and Conclusions.

## II. A BRIEF REVIEW OF WAVELETS

Before introducing the concept of a Haar Scattering Network, it is worth explaining what Wavelet Transforms and Haar Transforms are.

Readers should first notice that the main idea of these transforms is to decompose a signal into several different scale components. That said, a Wavelet transform is a transform, like Fourier, where its basis is composed by a family of orthonormal functions $\psi$, allowing to capture both frequency and location (in time and space), differently from the classical Fourier analysis.

In general terms, a Wavelet transform, as can be seen in [11], is defined by:

$$\chi_\omega(k,n) = 2^{\left(\frac{n}{2}\right)} \int_{-\infty}^{\infty} x(t) \Psi(2^n t - k) dt \quad (1)$$

$$\chi_\omega(k,n) = 2^{\left(\frac{n}{2}\right)} \int_{-\infty}^{\infty} x(t)\psi(2^n t - k)dt \qquad (2)$$

where they obey the following properties:

$$\psi_{(n,k)}(t) = 2^{\left(\frac{n}{2}\right)} \psi(2^n t - k) \qquad (3)$$

$$\int_{-\infty}^{\infty} \psi_{(n,k)}(t)dt = 0 \qquad (4)$$

$$\int_{-\infty}^{\infty} \psi_{(n_1, k_1)}(t) \cdot \psi_{(n_2, k_2)}(t)\, dt = \delta_{n_1,n_2}\delta_{k_1,k_2} \qquad (5)$$

$$\int_{-\infty}^{\infty} \psi^2_{(n,k)}(t)dt = 1 \qquad (6)$$

Equations (1) and (2) define how the transform coefficients can be calculated, allowing us to reconstruct the original signal as:

$$x(t) = \sum_{n,k=\infty}^{\infty} \chi_\omega(k,n) \cdot \psi_{(n,k)}(t) \qquad (7)$$

Equations (3) to (6) define the admissibility conditions for a specific function be considered a valid basis within the Wavelet framework, as can be seen in [11], and $\delta_{(n,k)}$ defines a delta Kronecker function. A special case of these functions is the Haar Wavelet.

A Haar Wavelet is defined by a function $\psi$, as in (8).

$$\psi(t) = \begin{cases} 1, & \text{if } 0 < t \leq \frac{1}{2} \\ -1, & \text{if } \frac{1}{2} < t \leq 1 \\ 0, & \text{otherwise} \end{cases} \qquad (8)$$

Also, its scaling function is defined in (9), as:

$$\Psi(t) = \begin{cases} 1, & \text{if } 0 \leq t \leq 1 \\ 0, & \text{otherwise} \end{cases} \qquad (9)$$

From (1), (2), (8) and (9), it is possible to derive a pair of equations for calculating the coefficients of the Haar Wavelet Transform:

$$\chi_\omega(k,n) = 2^{-\frac{1}{2}}(\chi_\omega(2k, n+1) + \chi_\omega(2k+1, n+1)) \qquad (10)$$

$$\chi_\omega(k,n) = 2^{-\frac{1}{2}}(\chi_\omega(2k, n+1) - \chi_\omega(2k+1, n+1)) \qquad (11)$$

Respectively, the coefficients obtained in (10) provides local averages of the functions in a specified scale, while the coefficients obtained in (11) provides the details of function in different scales. Mathematically speaking, we are extracting geometric features that are covariant to translations in the signal, as can be seen in [10].

III. DESCRIPTION OF HAAR SCATTERING NETWORKS

A Haar Scattering Network, following [8] and [9], is characterized by the iterative calculation of a permutation invariant operator defined by (12).

$$(\alpha, \beta) \rightarrow (\alpha + \beta, |\alpha - \beta|) \qquad (12)$$

One should notice that, first, that the sum $\alpha + \beta$ is proportional to the average between both numbers and $\alpha - \beta$, in specific conditions, can be regarded as a Haar Transform coefficient, as can be seen in (10) and (11).

Having said that, a Haar Scattering Network was originally defined in [8] and [9] by a sequence of layers, which operates over an input positive $d$-dimensional signal $x \in (\mathbb{R}^d)^+$.

Keeping that in mind, the network layers are defined as a two-dimensional array $S_j x(n, q)$ with dimensions $2^{-j}d \cdot 2^j$, where $n$ is a node number and $q$ denotes a feature index.

Following this definition, $S_j$ is a permutation invariant operator that acts over a set of nodes calculated in the previous layer following (13) and (14).

$$S_{j+1}(n, 2q) \rightarrow S_j(a_n, 2q) + S_j(b_n, 2q) \qquad (13)$$

$$S_{j+1}(n, 2q+1) \rightarrow |S_j(a_n, 2q) - S_j(b_n, 2q)| \qquad (14)$$

These equations generalize the original permutation invariant operator defined in (12) for pairs of nodes indexed by rules $a_n$ and $b_n$, which work as maps of pairs that can be optimized according to features to be extracted. In addition to that, still according to [8] and [9], it is also important to define the following identity:

$$S_0(n, 0) \rightarrow x(n) \qquad (15)$$

Consequently, the idea of the Haar Scattering Network is to iteratively extract Wavelets coefficients of the signal, and apply a pointwise absolute value operator over them (this point is going to be later explained in the present paper). Nonetheless, if we take a look at (12), it is possible to verify to recover the maximum and the minimum of the values by means of the relations in (16) and (17), respectively.

$$max(\alpha, \beta) \rightarrow \frac{1}{2}(\alpha + \beta + |\alpha - \beta|) \qquad (16)$$

$$min(\alpha, \beta) \to \frac{1}{2}(\alpha + \beta - |\alpha - \beta|) \quad (17)$$

Hence, it is possible to reconstruct the whole previous layer values just by these linear combinations, if $\alpha$ and $\beta$ are real positive values.

The similarities between our work and the original ideas within [8-10] cease here. Instead of thinking out how to recover the original signal (which, by definition is strictly positive) by applying the properties discussed in (16) and (17), our key insight is to work with the nonlinear properties of the pointwise absolute value operator together with possible signal reconstruction by means of Ordinary Least Squares and the contractive/scaling properties of Wavelets. As [8-10] have already demonstrated the powerful capabilities of this architecture as a feature extractor for classification purposes in a wide different set of problems (such as computer vision and others), in the present paper we only focus on its eventual capabilities as a function approximator.

To avoid any further and unnecessary complexity, suppose the signal $x(t)$ has a length equal to an arbitrary number that is power of two (this assumption simplifies the treatment with the dyadic operation). Also, suppose we want to map this function for each specific $t$, but this time, allowing $x \in \mathbb{R}^d$, in other words, $x(t)$ can exhibit negative real values. Thus, we are not able to reconstruct the signal using (16) and (17). Our approach is based on the projection of the geometric features extracted in an arbitrary layer $S_j(n, q)$.

Moreover, the pairing rules $a_n$ and $b_n$ are optimized in such way that we obtain a scale parameter $\sigma$ and a shift parameter $\tau$ acting over a signal of length $N$, where $a_n = n$ and $b_n = 2^{-j+1}$. $N \cdot \sigma + \tau + n$ that minimizes the distance between the reconstructed signal and a subsampling of the original signal.

The main idea of these pairing rules is that, instead of applying the traditional Haar filtering scheme in sub-sequent signal observations (i.e. $x(t)$ and $x(t-1)$), we treat the 1D signal as an entity that can be represented as kind of temporal graph – that must be identified – where each node represents a system state being connected due to their respective multiscale geometric features, such as spectral or frequency properties, which provide relevant information, motivated by the discussions carried out in [12], [13] and [14].

Another possible interpretation is the generalization of the traditional filtering operations, in contrast to those carried out in the traditional wavelet framework, in order to obtain the fundamental frequency that contains relevant information that enables us to understand invariants, symmetries and possible diffeomorphisms, as discussed in [12].

Following this scheme, our key insight is to calculate an Ordinary Least Squares estimate of the coefficients $\beta_n$, with $n \in [0, 2^j]$, that minimize the distance between $x(t)$ and $\hat{x}(t)$, where this last variable is the approximation of $x(t)$ using the geometric features extracted in $S_j(n, q)$. It is also important to notice that other classification/regression schemes can be applied, in substitution to the OLS.

That said, let $\hat{x}(t)$ be written as as a function of coefficients $\beta_n$, and geometric features extracted in an arbitrary layer $S_j$:

$$\hat{x}((n-1), 2^j + 1) = \sum_{k=0}^{2^j} \beta_k \cdot ((n-1) \cdot 2^j + 1) \cdot S_j(n, k) \quad (18)$$

In (18), it is possible to realize that the factor $((n-1) \cdot 2^j + 1)$ acts as a scaling factor that maps each level to the original function domain, and $\beta_k$ normalizes the feature and recovers the sign of each feature in terms of the original structure mapping. This approach is inspired by the reconstruction of the signal in terms of the Discrete Wavelet Transform coefficients, which is given by (19) and (20).

$$y_n = H_n x_n \quad (19)$$

where $x_n$ is the input signal, $y_n$ are the coefficients of the Haar Wavelet transform and $H_n$ is a linear operator that applies a Haar Wavelet Transform over $x_n$. Then:

$$x_n = H_n^{-1} y_n \quad (20)$$

In order to understand the scaling factor defined in (18), one must realize that the cascade of operations implicit in the recursive equations (13) and (14) can be defined by a cascade of Haar operators $H_n$, as in (21).

$$S_{j+1}(n, 2q) \to |H_j S_j(n, q)| \quad (21)$$

Every time $H_j$ acts over $S_j(n, q)$, the number of rows in the new matrix is half of the previous layer, following the definition of the Haar Wavelet Transform (see [11]). So, if the features used are being extracted at the layer $S_j$, our idea is to remap these features to the original domain set by interpreting that each line of the layer is a local feature related to each neighborhood. That said, the factor $((n-1) \cdot 2^j + 1)$ is a renormalization and translation factor that remaps each wavelet to a specific value in the domain set of the signal, given that $t \in [1, 2^d]$.

Therefore, (18) is basically derived as an adaptation of (20), taking into account that we apply a pointwise absolute value operator over the Haar Wavelet Coefficients, preventing us from directly reconstructing the original signal by inverting the Haar Matrix $(H_j)$. Now, we are able to reconstruct and interpolate the signal following the same implied geometric features.

Nonetheless, on top of (18), we can extend this idea to map a parameter that affects the data generation process of the sampled signal in a way that the signal is actually given by $x(t, \theta)$, or to identify the domain set (time) using the sampled data. This is where the nonlinear characteristic of the pointwise absolute value operator plays an essential role.

The first point is to calculate the average of the coefficients at each realization of the system given the parameter $\theta$. The second step is to map each average feature to the average point of the sampled data, which is $x(\frac{N}{2}, \theta)$, since we are averaging all features at different frequencies. After that, using a suitable transfer function, we can calculate an Ordinary Least Squares estimate in (22) as we previously did in (18).

$$\hat{x}\left(\frac{N}{2}, \theta_i\right) = \sum_{k=0}^{2^j} \beta_k \cdot \theta_i \cdot f(S_k(i,k)) \quad (22)$$

In (22), f is an arbitrary transfer function, $\theta_i \in [\theta_0, \theta_0 + i\tau]$ is an arbitrary parameter, which varies according to a step of size $\tau$.

It is also important to notice that pointwise absolute value operator turns the wavelet transform coefficients invariant, which is a desirable feature while calculating the average of the features in for each $\theta_i$. To better understand this statement it is important to remind that, by definition (3), $\psi_{n,k}(t-d) = \psi_{n,k+d}(t)$. In other words, Wavelets are covariant to translations. Introducing nonlinearities in the wavelet coefficients allows us to build invariant representations, as seen in [10] and [12], aiming to avoid the curse of dimensionality. We shall explain it below, by following the explanations described in [12].

While extracting the features that may describe the whole signal aiming to obtain a function $\widehat{f(b)}$ that approximates the true $f(b)$, where $b$ is the feature set and $f$ a function that maps these features to the temporal evolution or a specific class (such as a specific family of signals), a cascade of operations must be carried out to extract relevant spectral information. To circumvent this issue, it is desirable to define a contractive operator $\Phi(b)$ which reduces the range of variations of $b$, while still separating different values of $f$, in such way that $\Phi(b) \neq \Phi(b')$ if $f(b) \neq f(b')$. Our ultimate goal is obtaining a low dimensional vector $\Phi(b)$, where $f_0(\Phi(b)) = f(b)$. In this sense, it is said that $\Phi$ separates $f$. It is also worth noticing that (18) is its respective special linear case.

When the pointwise absolute value operator is introduced, the identity $\psi_{n,k}(t-d) = \psi_{n,k+d}(t)$ no longer holds, enabling us to find $\sigma$ and $\tau$ in a way that it allows to identify directions which $f(b)$ does not vary, i.e. what are the translations in the time series (in the 1D case) that the features do not vary, pointing out to possible symmetries in time, in each layer $S_j$, as can be seen in Fig. 1, for a sinusoidal wave. When we remove the pointwise absolute value operator, we obtain Fig. 2, for the same $\sigma$ and $\tau$ (which were not optimized).

Features in Fig. 1 retain significant symmetry and other information about the signal (e.g. symmetries around 300, 2300, 4300, 6300 and 8300, where maxima and minima occurs; and changes of signal as nearby point 1300, for example), while features in Fig. 2 only reproduce the average cyclical components present in the original signal.

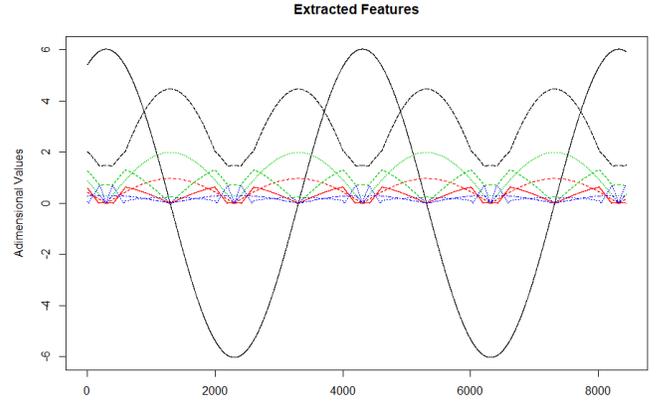

Fig. 1. Extracted Features of a Sinusoidal Signal using a three-layer Haar Scattering Network containing each feature $q$.

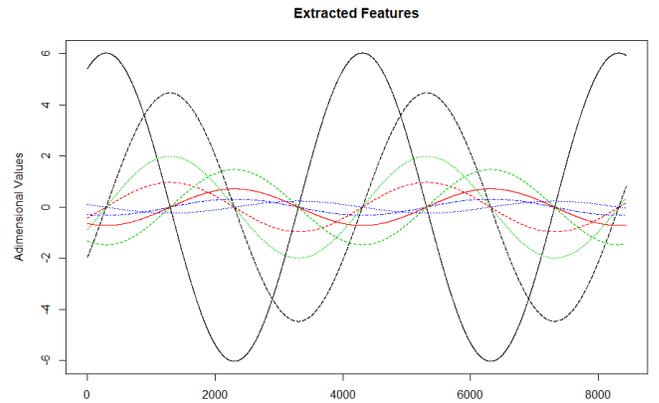

Fig. 2. Extracted Features of a Sinusoidal Signal using a three-layer Haar Scattering Network, without the pointwise absolute value operator, containing each feature $q$.

In addition to that, given the fact that the layers are calculated with a cascade of operations, it is desirable that $\|H_j S_j(n,q)x - H_j S_j(n,q)x'\| \geq \|S_{j+1}x - S_{j+1}x'\|$, otherwise, as soon as we calculate each layer, the values may diverge to $\pm\infty$. The pointwise absolute value ensures that this divergent behavior will not happen, as seen in [12] and [8].

As consequence, we can extract important spectral/frequency information that can be used to perform the desired tasks.

Thus, having presented our approach, we test it under a set of different problems and see how our network performs.

## IV. COMPUTATIONAL EXAMPLES

In the present section we show that Haar Scattering Networks can be used to decompose several different functions into its geometric features and reconstruct them. To accomplish that, four different computational exercises were made:

decomposition and reconstruction of sinusoidal signals; decomposition and reconstruction of exponential signals; decomposition and reconstruction of a non-linear model; and, finally, identification of an autoregressive parameter.

With these simple exercises, we hope showing a new direction towards the construction of bijective functions that can be used to detect frequencies, identify stochastic parameters on linear and non-linear systems, make predictions and possibly other applications beyond the original ones, which basically aimed computer vision and classification problems.

Keeping that in mind, we expect to demonstrate its potential applications as a general-purpose non-linear regression tool built on top of very simple computational operations that can be even calculated using spreadsheet software.

The computational exercises consist of: Simulate the processes; Extract the features using a four-layer Haar Scattering Network; and regress the extracted features against systems states, parameters or time scales (using eqs. 18 or 22).

All computations can be provided upon request to the author.

A. *Decomposition and Reconstruction of Sinusoidal Signal:* Aiming to demonstrate the capabilities of this tool to decompose and reconstruct signals, our first example demonstrates what the output of a four-layer Haar Scattering Network acting over $x(t) = \sin((\beta \cdot 2\pi \cdot t)/3600)$ is, where $t \in [1; 1024]$, in order to facilitate the dyadic cascade of operations (as we have 210 samples). We have extracted average geometric features for $\beta \in [-6; 6]$ in Fig. 3, following (13) and (14).

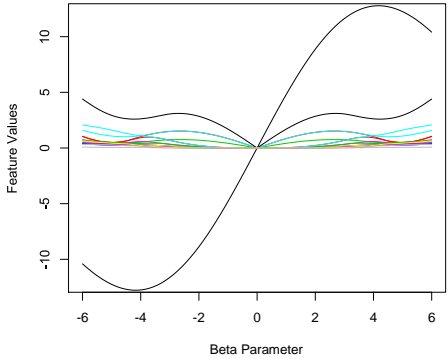

Fig. 3. Extracted Features of a Sinusoidal Signal, according to the $\beta$ parameter, using a four-layer Haar Scattering Network, where each color represents a feature $q$.

While reconstructing the signal, we have obtained a $R^2$ measure of 99 percent, using only 4 layers. Hence we are able to reconstruct the function within any $\beta \in [-6; 6]$.

B. *Decomposition and Reconstruction of an Exponential Signal:* In this second exercise, we have repeated the same simulation scheme, but using an exponential function of the form $x(t) = \exp(\beta \cdot \frac{t}{3600})$.

Following the same scheme, we also have obtained a $R^2$ measure of 99 percent, using only 4 layers.

C. *Decomposition and Reconstruction of a Nonlinear Noisy Signal:* Both previous exercises relied on simple and deterministic signals. Now, we are going to evaluate a non-linear difference equation inspired in a simple population model in which a saturation behavior is introduced, presented in (23).

$$x(t) = x(t-1) + \frac{\beta}{C} \cdot x(t-1) \cdot [C - x(t-1)] + \epsilon(t) \quad (23)$$

In (23), $\beta$ denotes the population growth factor, $C$ denotes the carry capacity of the system, $\epsilon(t)$ denotes a perturbation in the system, which might be caused by deaths/births related to exogenous factors such as diseases, birth policies or any other kind of phenomena, and $x(t)$ denotes the population level itself.

That said, we test how the Haar Network performs in the approximation of the simulated signal (for forecasting purposes), and how it performs by calculating the systems state at any arbitrary instant.

In the case of the system estimation of the simulated signal as a time series procedure, we have obtained a $R^2$ measure of 96.11 percent (which can be seen in Fig. 4), for $\beta = 0.005$.

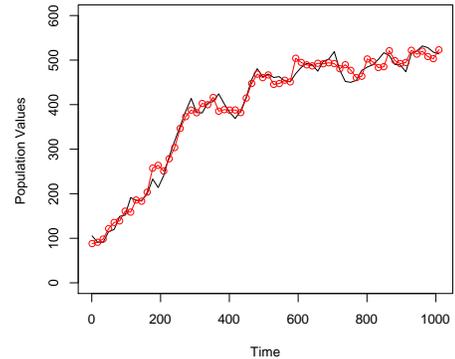

Fig. 4. Approximation of the simulated nonlinear signal. Red dots represent estimated values, while the black line represent the real values.

In the case of approximating the system's state at any arbitrary instant, given a $\beta$ parameter, the obtained $R^2$ measure was 91.5 percent.

D. *Identification of the Autoregressive Parameter:* Now, suppose we have an Autoregressive model as in (24).

$$x(t) = \phi \cdot x(t-1) + \epsilon(t) \qquad (24)$$

Also, suppose that we have two ways to estimate the parameter $\phi$. One way is introducing a unit step in the system, which represents a constant energy input, allowing us to verify how the system goes toward a new equilibrium point.

Another way is to introduce a pulse function in the system, which is a way to verify how the energy is dissipated in the system. Usually these techniques are applied to extract the deterministic part of the system evolution from the Gaussian disturbances $\epsilon(t)$ by means of convolution analysis or autocorrelation function analysis.

Following the same scheme of the previous example, our idea is to extract the geometric/mathematical features of the process when it receives an input and, following (22), map the system evolution to a specific parameter.

By applying these steps, we obtain a $R^2$ measure of 96.63 percent, that can be verified in Fig. 5, while trying to map the system simulation to each parameter $\phi$.

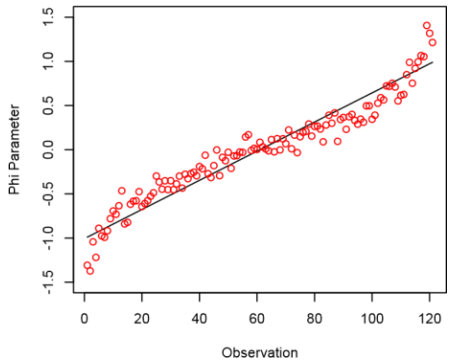

Fig. 5. Inference of parameter $\varphi$. Red dots are estimated parameter values and the black line represents the real values.

## V. DISCUSSION OF THE RESULTS

Given all the different simulations, it is possible to see that this kind of network can enhance the comprehension of several phenomena, in terms of classification and approximation problems, extending its original purpose, which was feature extraction of images and graphs, to classification problems. For example, for sinusoidal signals, the features seems to represent details for each mapped parameter. The same for exponential signals, and so on. Hence, basically, the extracted features actually have an understandable mathematical interpretation.

When $R^2$ measure is calculated for each one of these examples, using only four computational layers plus a linear regression structure, all examples have $R^2$ greater than 90 percent.

It is also possible to verify that this architecture has a very low computational cost since it is built over simple algebraic operators, such as subtractions, additions and absolute value operators. Thus, these computations can be made in any kind of mathematical software available to the final user, and can be very easily implemented.

In addition to that, the most important aspect of this kind of network architecture is that it enables us to build very simple regression structures on top of the extracted features. For all problems here investigated, only linear operations were carried out.

Finally, it is important to notice that the signal extraction and feature mapping can be extended to multiple parameters (and variables) by means of building multivariate quasi-bijective functions, where each set of parameters (variables) is mapped into a unique real number, which then can be regressed against the extracted features.

## VI. CONCLUSIONS

Along this paper it was discussed the possibility of building a general purpose approximator on top of Haar Scattering Networks. Moreover, it was pointed out that there are huge possibilities such as identification of stochastic parameters on linear and non-linear systems, forecasting and classification problems and possibly other applications beyond the original ones, which basically aimed computer vision and classification problems in their respective implementations, as can be seen in [13] and [14].

Exploring the fact that Wavelets allow us to capture relevant multiscale information, but being covariant to translations, when nonlinearities are introduced in the wavelet coefficients calculation, we are able to build invariant representations, where we can build deep networks to retrieve an extensive amount of features with desirable properties in a stable way, allowing simpler regression structures, but still, being able to capture nonlinear features of the dataset, which were shown in the computational examples. In other words, it is possible to obtain more human-understandable machine learning structures.

Given these interesting results, we think there is a huge list of tasks to be done, as a perspective for future works. First, it would be interesting to compare the performance of the architecture presented here with most traditional tools. Also, it would be very interesting to check on which conditions this kind of network performs better or worse. And, finally, it is very interesting to check other potential applications that were not considered here, such as time series classification problems, forecasting problems and so on.

### ACKNOWLEDGMENT

We would like to thank Pedro Delano Cavalcanti for able research assistance. We would also like to thank Prof. Claudio Garcia, at Polytechnic School of Engineering (University of São Paulo) and Prof. Rodrigo de Losso da Silveira Bueno at School of Economics, Business and Accounting (University of São Paulo) for suggestions, comments and discussions.

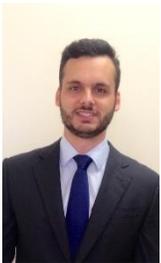

**Fernando Fernandes Neto** was born in Taboão da Serra, Brazil, in 1986. He completed his PhD degree in Systems Engineering in 2016 at Polytechnic School of Engineering – University of São Paulo, Brazil. Earned his MSc degree in Industrial Engineering at São Paulo Institute of Technology (IPT) in 2014, MBA in Financial Engineering at University of São Paulo in 2013 and received his BSc in Business Administration in 2010.

Currently, he is a post-doctorate fellow at University of São Paulo, researching spectral methods and machine learning methods applied to systems identification, econometrics and forecasting.